\documentclass[acmlarge,nonacm]{acmart}

\usepackage{subcaption}  % for subfigures

\AtBeginDocument{%
  \providecommand\BibTeX{{%
    \normalfont B\kern-0.5em{\scshape i\kern-0.25em b}\kern-0.8em\TeX}}}

\begin{document}

\title{Atunda: Building an Afrogenic Dance Movement Dataset for AI-Driven Classification}

% \title{The Name of the Title is Hope}

\author{Diego Carrascosa}%
\authornote{The first two authors contributed equally to this research.}
\email{diego.carrascosa@ufl.edu}%diegoj.carrascosa@gmail.com
\author{Tristan Pank}%Tristan Pank
\authornotemark[1]
\email{tristan.pank@ufl.edu}%panktristan@gmail.com
\affiliation{%
 \department{Digital Worlds Institute}
  \institution{University of Florida}
  %% \streetaddress{P.O. Box 1212}
  \city{Gainesville}
  \state{FL}
  \country{USA}
  \postcode{32611}
}

\author{Qudus Onikeku}%Qudus Onikeku
\affiliation{%
  \institution{QDance Center}%QDance Center
  % \streetaddress{P.O. Box 1212}
  \city{Lagos}
  \country{Nigeria}
}
\email{qudus@qdancecenter.com}%qonikeku@arts.ufl.edu

\author{Mas'ud Onikeku}%Qudus Onikeku
\affiliation{%
  \institution{QDance Center}%QDance Center
  % \streetaddress{P.O. Box 1212}
  \city{Lagos}
  \country{Nigeria}
}
\email{masudonikeku@gmail.com}%qonikeku@arts.ufl.edu

\author{Jeremy A Frusco}%Jeremy A Frusco
\affiliation{%
  \institution{University of Florida}
  %% \streetaddress{P.O. Box 1212}
  %\city{Gainesville}
  %\state{FL}
  \country{USA}
  %\postcode{32611}
}
\email{jfrusco@ufl.edu}%jfrusco@ufl.edu

\author{O\d{s}ubi Craig}%O\d{s}ubi Craig
\affiliation{%
\institution{University of Florida}
  %% \streetaddress{P.O. Box 1212}
  %\city{Gainesville}
  %\state{FL}
  \country{USA}
  %\postcode{32611}
}
\email{osubi.craig@ufl.edu}%ocraig@arts.ufl.edu

\author{Angelos Barmpoutis}%
\affiliation{%
\institution{University of Florida}
  %% \streetaddress{P.O. Box 1212}
  %\city{Gainesville}
  %\state{FL}
  \country{USA}
  %\postcode{32611}
}
\email{angelos@digitalworlds.ufl.edu}%

%\renewcommand{\shortauthors}{Anonymous Author 1 and 2, et al.}
 \renewcommand{\shortauthors}{Carrascosa and Pank, et al.}

\begin{abstract}
  This paper introduces Atunda, the first systematically collected video dataset of Afrogenic dance moves, designed for training and evaluating deep learning models for movement classification. The first phase of the dataset includes nearly 400 videos covering 13 distinct dance moves from African and African Diaspora performative culture, making it the first AI-ready resource of its kind. Using this dataset, we train and evaluate various configurations of temporal convolutional neural networks, achieving a classification accuracy exceeding 95\%. The baseline results provide a benchmark for future studies in movement classification, especially in the understudied area of Afrogenic performative culture and tradition. This work lays the foundation for developing in the future digital metrics of similarity, automated attribution systems, and tools that preserve and promote the cultural heritage of Afrogenic dances.
\end{abstract}

\maketitle

\section{Introduction}
\label{sec:introduction}

Dance has been a popular form of creative expression across cultures that has been carried from generation to generation through oral tradition. The importance of documenting and preserving such cultural dances as part of the world's Intangible Cultural Heritage (ICH) has been recognized by the UNESCO 2003 convention \cite{unescoICH}. However, in addition to traditional dances, contemporary dance routines have become a well-recognized form of pop culture that is globally disseminated through social media platforms. Through this large-scale creative process, dance moves can become viral, massively reproduced, evolved, and transferred to other media such as video games. These moves can then become a source of profit, as in the case of emotes---short, animated expressions or gestures often used in video games---which can be bought as in-game purchases, generating significant revenue for software development companies, often without crediting the original creators.

The influence of technology on pop-culture has created new challenges and possibilities, opening new areas for scholarly research. How can we document the evolution of a popular dance routine? How can we properly attribute a viral dance move to the original creator? Subsequently, can dance as an artistic creation be protected by copyright laws? The key motivation of this paper, and the broader vision behind our work, is to leverage artificial intelligence (AI) methods to create tools for identifying dance moves and facilitating their attribution, with a focus on Afrogenic dance moves.

According to the literature, there is little protection for ordinary choreographers of dances on social media, and no protection for choreographers’ works that are considered "social dances" or consist of only a few steps, as they are regarded as simple blocks of artistic expression \cite{pinkston2022}. In 2020, JaQuel Knight, an American choreographer and dancer, made a strong push for dance copyrights by encoding his work "Single Ladies" in Labanotation, making it one of the first non-ballet choreographic scores to be approved for copyright \cite{knight_forbes}. This demonstrated that copyrighting a dance routine is possible if it consists of more than just a few steps and is sufficiently documented in a form of notation that can serve as legal evidence.

Labanotation or kinetography Laban is a notation system developed by Rudolf Laban for recording movement \cite{Laban1966,Hutchinson1977,longstaff2001}. Laban Movement Analysis (LMA) has been systematically studied \cite{Bernardet2019}, further developed, and used for encoding movement in a variety of applications involving the study of human motion, such as dance movement therapy, art performances, non-verbal body-based communications \cite{roudpoahti2012,davies2007}, and clinical tests \cite{Foroud2006}. Other notable examples include LMA visualization systems for dance instruction and assessment \cite{Rett2010} as well as for supporting the inclusion of wheelchair dancers \cite{xie2023movement}. 

Several computer-based LMA methods have been developed for the automatic generation of Labanotation from motion capture data, aiming to streamline the labor-intensive process of manual notation \cite{li2019dance,cui2021method}. The advancements on motion capturing technologies and their availability as low-cost consumer devices, such as the series of Kinect sensors by Microsoft (2011-2023), have contributed significantly to this field. Several Kinect-based methods for LMA have been proposed \cite{Santos2011,Bernstein2015}, with various applications, such as emotional motion recognition \cite{Kim2013}, Labanotation instruction \cite{raheb2021moving}, automated dance evaluation \cite{jang2017automated}, and others.

Beyond Labanotation, several machine learning algorithms, such as Random Decision Forest, Multi-Layer Perceptron and Multi-class Support Vector Machines have been used for analyzing motion data captured from Kinect or other motion tracking sensors \cite{Ajili2017,Guo2022,jin2022detecting}. Deep learning methods such as convolutional neural networks with long short-term memory have been used for dance emotion recognition \cite{Wang2020} and classification of choreographic primitives \cite{bakalos2020adaptive}. Sun et al. \cite{sun2015human} proposed a 3D spatio-temporal convolutional neural network (TCN) architecture. This method directly inputs videos into the TCN, utilizing both the spatial aspect of pixels and the temporal aspect of continuous video frames for classification.  Similarly, other types of deep learning architectures, such as Visual Geometry Group Networks, have been used for movement recognition, providing real-time feedback in computer-assisted dance education \cite{jin2022fusion,kang2023dancing}. 

For an in-depth overview of techniques for capturing and storing dance-related content, as well as methods for extracting representative primitive patterns, recognizing choreographic poses and movements, analyzing choreographic patterns, and summarizing or evaluating performances, the reader is referred to recent literature reviews \cite{kico2018digitization,rallis2020machine,protopapadakis2020digitizing,hou2022digitizing,reshma2023cultural,Ontology} that discuss the strengths and limitations of the state-of-the-art methods.

This technological revolution has resulted in many useful end-user applications, such as real-time feedback for dance learning and training \cite{li2023research,rallis2021bidirectional,bakalos2020motion,raheb2019dance}, dance summarization \cite{voulodimos2018kinematics,rallis2018spatio,rallis2017extraction}, virtual and augmented reality applications for cultural and social dancing  \cite{pistola2021creating,sarupuri2023dancing, bozia2018reviving}, and 
AI-generated choreographies \cite{papillon2023pirounet,sung2020deepdance}. Several of these applications have been showcased across a wide variety of genres, including, among others, Cantonese opera \cite{chung2023utilising}, Indian classical dance \cite{parthasarathy2023novel,dewan2018spatio} including Bharatanatyam dance\cite{Bharatanatyam2022,Bharatanatyam2025,Bharatanatyam2023}, Lazgi dance \cite{skublewska2021methodology}, indigenous Aeta dances \cite{mindoro2021comparative},  Slovak folk dances \cite{hajdin2019digitization}, Greek folk dances \cite{Aristidou2015}, and martial arts demonstrations \cite{yamanaka2021evaluation}. This multitude of studies has led to the development of open-access movement databases, some of which are specific to particular genre or culture  \cite{UniversityCyprus2018}, and others that are multi-cultural such as the TikTok dance move dataset that was primarily used for depth estimation of humans \cite{Jafarian_tiktok} and the AIST Dance Video Database, which focuses on 10 general and well known dance genres, and includes multiple performers, and cameras \cite{tsuchida2019aist}.  Finally, there has been significant number of annotation tools \cite{elraheb2021annotating}, and open-source pose tracking libraries such as SAM-Body4D, FMPose3D, MediaPipe, Yolo8, Detectron2, and others, which are important to further facilitate research in this field \cite{gao2025sambody4d,yang2026sam3dbody,wang2026fmpose3dmonocular3dpose,NguyenMediapipe,Wu2019,yolo8}.

Despite the plethora of publications on this broad topic of research, there are not many studies with focus on Afrogenic dance moves or any systematically recorded datasets to facilitate research on African dance culture. Furthermore, there is a significant gap in the literature regarding computer-assisted mechanisms for attributing dance moves to their creators. Identifying solutions to this problem will have a major impact, particularly on underrepresented dance creators, who often go unaccredited despite their dance moves becoming viral on social media and being monetized in video games \cite{pinkston2022}.

The work presented in this paper is part of our broader project vision to develop a deep learning model that can successfully classify different dance movements, laying the foundation for future development of digital metrics of similarity. These metrics could be used to track the evolution of dance moves, provide attribution, and help protect dance artists' creative work. By using deep learning to classify movements, the model and its predictions can serve as evidence of ownership for digital assets, such as digitally encoded dance routines, thereby helping creators receive their deserved attribution.

In this paper, we introduce a novel video dataset consisting of 13 different African and African Diaspora (Afrogenic) dance moves, performed in almost 400 videos. This dataset was created through Project Atunda, an interdisciplinary research initiative involving a team of artists, scientists, and researchers who collaborated to build an AI-ready video data bank of Afrogenic dances. The dataset created in this project was used to train and test a preliminary deep learning model based on a temporal convolutional neural network. The results are discussed in detail and serve as a benchmark for assessing other dance move classification methods in the future.

The key contributions of this paper are as follows: 1) We present Atunda, a novel video dataset of Afrogenic dance moves that has been systematically collected and curated for training machine learning models. To the best of our knowledge, this is the first dataset of its kind and aims to promote and facilitate research on Afrogenic dance and culture. 2) We present results from training and validating various configurations of temporal convolutional neural networks using the Atunda dataset. The results demonstrate classification accuracy exceeding 90\% across all 13 types of dance moves in our dataset and can be used as a baseline benchmark for future studies utilizing the Atunda dataset.

\section{Dataset Development}
This section presents Atunda, a video dataset of Afrogenic dance moves  created specifically for this project. The term "Atunda" has its roots in the Yoruba language, which is widely spoken in Nigeria, particularly in the southwestern region. In Yoruba, "Atunda" is often associated with the idea of renewal or rebirth. The concept behind "Atunda" is often linked to the idea of starting afresh, creating something new out of something old, or experiencing a form of renaissance. The purpose of the Atunda dataset is to provide researchers in the field of human action recognition with a unique set of movements typical of Afrogenic dance, which can be used for machine learning training and testing. Contrary to other dance movement datasets discussed in Sec. \ref{sec:introduction}, which include samples from multiple genres, such as the TikTok \cite{Jafarian_tiktok} and AIST \cite{tsuchida2019aist} datasets, our Atunda dataset provides a focused, Afrogenic-specific set of movements capturing diverse dance practices, ranging from heritage-inspired forms to modern street and popular Afrobeats styles.

Afrogenic dance movements encompass a broad spectrum of expressive practices shaped by diverse historical, cultural, and technological contexts. The dance moves included in the Atunda dataset reflect this variability and can be broadly grouped into three overlapping categories: heritage-derived and ceremonial forms, popular music-driven choreographies, and contemporary street and social media trends. This categorization highlights the range of transmission pathways through which African dance vocabularies evolve, from traditional communal performance and ritual practice to commercial music circulation and digital platform dissemination. Although these categories are not rigid, they provide a useful framework for understanding the sociocultural dimensions that influence the movement structures represented in the dataset.

The first category includes movements rooted in cultural heritage and ceremonial performance, such as Akwaaba, Omo Iya, and Bante. These dances are closely tied to communal identity, ritual symbolism, and intergenerational cultural transmission. Akwaaba, (meaning “welcome” in the Twi language of Ghana’s Akan people) represents a welcoming, high-energy West African dance and drumming tradition rooted in Akan culture, commonly performed at festivals, formal receptions, and state-sponsored cultural events, where it expresses hospitality and reinforces collective identity \cite{akwaaba}. Omo Iya, grounded in Yoruba cultural traditions, reflects themes of kinship, ancestry, and communal celebration, with connections to ceremonial forms such as Gelede that honor feminine power and social harmony. Similarly, Bante is associated with Yoruba charm-related practices. In these cases, dance extends beyond entertainment to serve ritual, social, and at times political purposes through the preservation and public performance of cultural identity.

The second category consists of dances popularized through commercial music and mass media, including Alanta, Etighi, and Chocobodi. These movements illustrate the role of the African music industry in transforming localized dance practices into nationally and internationally recognized choreographic forms. Alanta, which emerged in Nigeria in 2007 through Artquake’s music video, became widely recognized for its theatrical hand gestures and exaggerated facial expressions \cite{Alanta}. Etighi, originally associated with the Ibibio and Efik peoples of Akwa Ibom, gained widespread visibility through its adaptation in Nigerian popular music videos, particularly through artist-driven dissemination \cite{etighi}. Chocobodi, popularized by A-Star, represents a newer Afrobeats-associated dance style characterized by energetic structured movement combined with freestyle interpretation \cite{chocobodi}. These dances occupy a sociocultural space where performance, entertainment, and media circulation intersect, reflecting how contemporary African popular culture reshapes traditional and emerging movement vocabularies.

The third category includes contemporary urban street and digitally mediated dance trends such as Gbeku, Gwara, Jazze, and Swo. These forms are primarily shaped by youth culture, social interaction, and online participatory platforms such as TikTok and Instagram. Gbeku, associated with Nigeria’s Zanku movement and artists such as Zlatan, emphasizes relaxed rhythmic grooves and reflects the creativity of Lagos street dance culture \cite{gbeku}. Gwara, derived from Southern African street dance traditions, was created by disc jockey and producer DJ Bongz, and was heavily performed and popularized in the mid-2010s \cite{gwara}, while Jazze represents Gabonese urban Afro-dance emphasizing grounded movement and body isolations. Similarly, Swo exemplifies short-form social media choreography designed for rapid replication and adaptation. These movements are characterized by fluid evolution, peer-driven innovation, and transnational digital dissemination, demonstrating how African dance continues to adapt to contemporary social and technological environments.

The Atunda dataset videos are systematically collected by the QDance Center \cite{qdance}. The QDance Center is a creative incubator based in Lagos, Nigeria, that leverages artistic competence for sociocultural development. The center is at the forefront of showcasing Nigerian culture to the world through dance, music, and cultural engagements. Its mission follows a six-pronged approach known as QDance CREATE, focusing on Community, Research, Education, Artistic excellence, Technocultural advancements, and Entrepreneurship. The QDance Center is the custodian of the Atunda video dataset and is responsible for the curation, and distribution of the videos, while the performers retain the ownership of their artwork.

The development of the Atunda dataset follows a three-phase strategy designed to ensure scalability, data quality, and broad community participation. In the first phase, the results of which are presented in this paper, a pilot dataset was created consisting of 13 dance moves and approximately 400 videos, all performed by dancers from the QDance Center. The objective of this phase was to establish a baseline dataset and identify effective recording and annotation practices that will guide subsequent data collection. During this phase, all labels were assigned by expert members of the QDance Center to ensure high annotation quality. In the second phase, the dataset will be expanded through a crowd-sourcing approach involving the wider network of the QDance Center, resulting in an intermediate dataset of approximately 50 dance moves and 1,000 videos. This phase focuses on the development of a dedicated crowd-sourcing platform \cite{atundaLive}, the establishment of curation and quality control criteria, and a hybrid labeling process in which labels are initially provided by the performers and subsequently verified by expert members of the QDance Center. Both the second and third phases are supported by sponsorship from the Robert Wood Johnson Foundation, which provides compensation to participating performers; verification of labels by QDance Center experts is required prior to payment approval. In the final phase, the dataset will be further extended to include up to 100 dance moves and approximately 2,000 videos, following the same crowd-sourcing and expert-validated labeling process as in the second phase. 

To ensure consistency and sufficient representation across all phases, the Atunda dataset follows a standardized data collection protocol. Specifically, each dance move is performed by at least five different performers and is represented by a minimum of 15 videos. To facilitate scalability and ease of contribution, all videos adhere to a simple and reproducible recording setup inspired by social media practices: monocular, portrait-oriented recordings that capture the full body of the dancer in a front-facing view, performed in everyday clothing and natural, non-controlled environments.

\begin{figure}[h]
    \centering
    \includegraphics[width=1\textwidth]{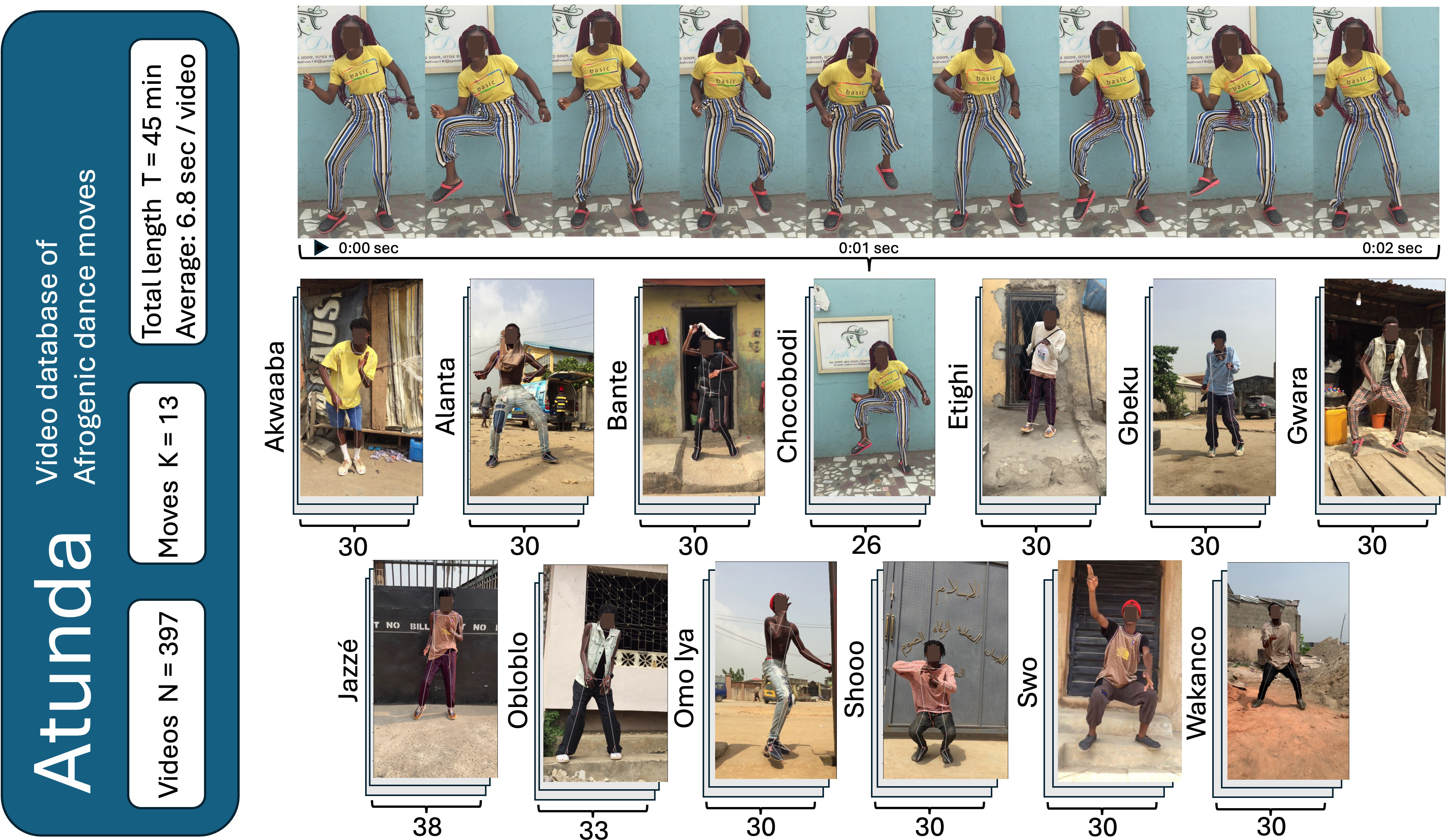}
    \caption{The Atunda video dataset of Afrogenic dance moves}
    \label{fig:dataset}
\end{figure}

In this paper, we present the first phase of the Atunda dataset. Each video captures a specific dance move performed by a single dancer or a pair of dancers in outdoor urban settings in Lagos, Nigeria under sunny conditions, as shown in Fig. \ref{fig:dataset}.  This design choice follows the aforementioned established protocol and reflects real-world recording conditions while eliminating the need for specialized equipment, thereby enabling straightforward replication and supporting future dataset expansion through crowd-sourcing, as discussed in Sec. \ref{sec:future_work}.

The current dataset consists of 397 videos totaling 45 minutes and covering the following 13 movements: Gwara, Obloblo, Etighi, Wakanco, Chocobodi, Jazz\'{e}, Bante, Akwaaba, Gbeku, Shooo, Swo, Omo Iya, and Alanta. The dance moves were performed by 17 members of the QDance Center (16 male and 1 female), aged between 20 and 30 years, all with a minimum of five years of dance experience, who voluntarily agreed to be recorded for the purposes of this project. Each movement was performed 30 times by different dance artists in different outdoor environments and with different clothing, resulting in an equal number of videos, except for Bante with 26, Gwara with 38, and Obloblo with 33 videos. Depending on the type and complexity of the dance move, each performance lasts between 4 and 14 seconds. The majority of the videos in the dataset are single-person performances, with only a few featuring multiple dancers. All videos are in a 9:16 aspect ratio (vertical) and vary in resolution and frame rate, as detailed in Table ~\ref{tab:video_description}. The table reports, for each type of movement, the total number of videos, their vertical resolution, their frame rate, the total number of frames, and the total duration of the videos. 

\begin{table}[]
    \centering
    \caption{Video properties in the Atunda dataset. (*One Chocobodi video is 720p.)}
    \label{tab:video_description}
    \begin{tabular}{|c|c|c|c|c|c|c|c|}
        \hline
        \textbf{Move} &\textbf{Videos} & \textbf{1080p} & \textbf{2160p} & \textbf{30fps} & \textbf{60fps} & \textbf{Frames} & \textbf{Seconds}\\
        \hline
        Akwaaba & 30 & 30 & 0 & 30 & 0 & 5854 & 195.33\\
        Alanta & 30 & 30 & 0 & 30 & 0 & 4773 & 159.26\\
        Bante & 26 & 26 & 0 & 26 & 0 & 2895 & 96.60\\
        Chocobodi & 30 & 29* & 0 & 1 & 29 & 23399 & 408.23\\
        Etighi & 30 & 0 & 30 & 16 & 14 & 7949 & 179.13\\
        Gbeku & 30 & 30 & 0 & 30 & 0 & 5351 & 178.55\\
        Gwara & 38 & 38 & 0 & 27 & 11 & 8088 & 205.32\\
        Jazz\'e & 30 & 30 & 0 & 0 & 30 & 15630 & 260.50\\
        Obloblo & 33 & 33 & 0 & 33 & 0 & 10221 & 341.04\\
        Omo Iya & 30 & 30 & 0 & 30 & 0 & 4944 & 164.96\\
        Shooo & 30 & 30 & 0 & 30 & 0 & 5014 & 167.30\\
        Swo & 30 & 30 & 0 & 30 & 0 & 5233 & 174.61\\
        Wakanco & 30 & 1 & 29 & 30 & 0 & 4777 & 159.39\\
        \hline
        Total & 397 & 337 & 59 & 313 & 84 & 104128 & 2690.22\\
        \hline
    \end{tabular} 
\end{table}

Since the primary technological purpose of this dataset is to build dance move classifiers, it is important to note its limitations. First of all, not all dance moves are equally represented with the same amount of data. Although each dance move has a similar number of associated videos, the actual amount of data varies depending on the number of frames in each video. Table ~\ref{tab:video_description} shows the total number of frames for each type of dance move. It is evident that there are some imbalances in the dataset by type, which may require data balancing before training a machine learning model. Furthermore, the gender diversity is evidently limited in the preliminary dataset due to the limitations in our cohort. As we continue to expand this dataset during phases 2 and 3, we aim to collect data that will result in a more balanced representation and involve the broader community through crowd sourcing. 

Finally, as indicated in Table ~\ref{tab:video_description}, 84 of the 397 videos were captured in 60 fps, with the remaining 313 videos in 30 fps. Since the temporal dimension of the input to a machine learning model is represented as discrete frames, any differences in frame rate may cause the algorithm to perceive some dance moves as slower or faster than others. Although this could be easily mitigated by normalizing the frame rates, we chose to leave the input videos unaltered during data processing to prevent the model from developing a bias toward a specific frame rate or movement speed, thereby accommodating variability in creative expression. It should also be noted that the videos were recorded as standalone rhythmic movements rather than as part of choreographies accompanied by specific music; therefore, this dataset is not suitable for music–movement synchronization studies \cite{endo_music}.

\section{Data processing}
\label{sec:data_processing}
Access to the raw videos can be requested from the QDance Center \cite{qdance} by providing a data review protocol approved by the ethics review board of an academic institution. The University of Florida was the first research institution to partner with the QDance Center and gain access to the Atunda video dataset for dance move classification research under the IRB202500278 data review protocol, which was approved by the University of Florida Institutional Review Board. This section describes the data processing steps applied to the Atunda dataset, with two main goals: 1) to extract non-identifying descriptive features from each video and create a comprehensive metadata record to be included in the Atunda dataset, and 2) to train a preliminary machine learning model that can serve as a comparative reference or benchmark for future dance move classification studies using the Atunda dataset.

\subsection{Input feature extraction}
\label{sec:pose_estimation}

Pose-based methods are considered one of the most promising directions for human action recognition (HAR), as indicated in a recent literature survey by Kulbacki et al. \cite{Kulbacki:2023:IHA:37177461}. One reason pose-based HAR is promising is that pose estimation methods are abundant and provide enriched data that are highly descriptive and simpler to use for further training. The process of calculating pose-based input features for machine learning begins by extracting individual frames from each video and processing them with a skeleton fitting method such as MediaPipe Pose Detection library \cite{NguyenMediapipe}, FMPose3D \cite{wang2026fmpose3dmonocular3dpose}, Detectron2 \cite{Wu2019}, Yolo8 \cite{yolo8}, SAM-Body3D \cite{yang2026sam3dbody}, SAM-Body4D \cite{gao2025sambody4d} or other pose estimation algorithms. 

\begin{figure}[h]
    \centering
    \includegraphics[width=1\textwidth]{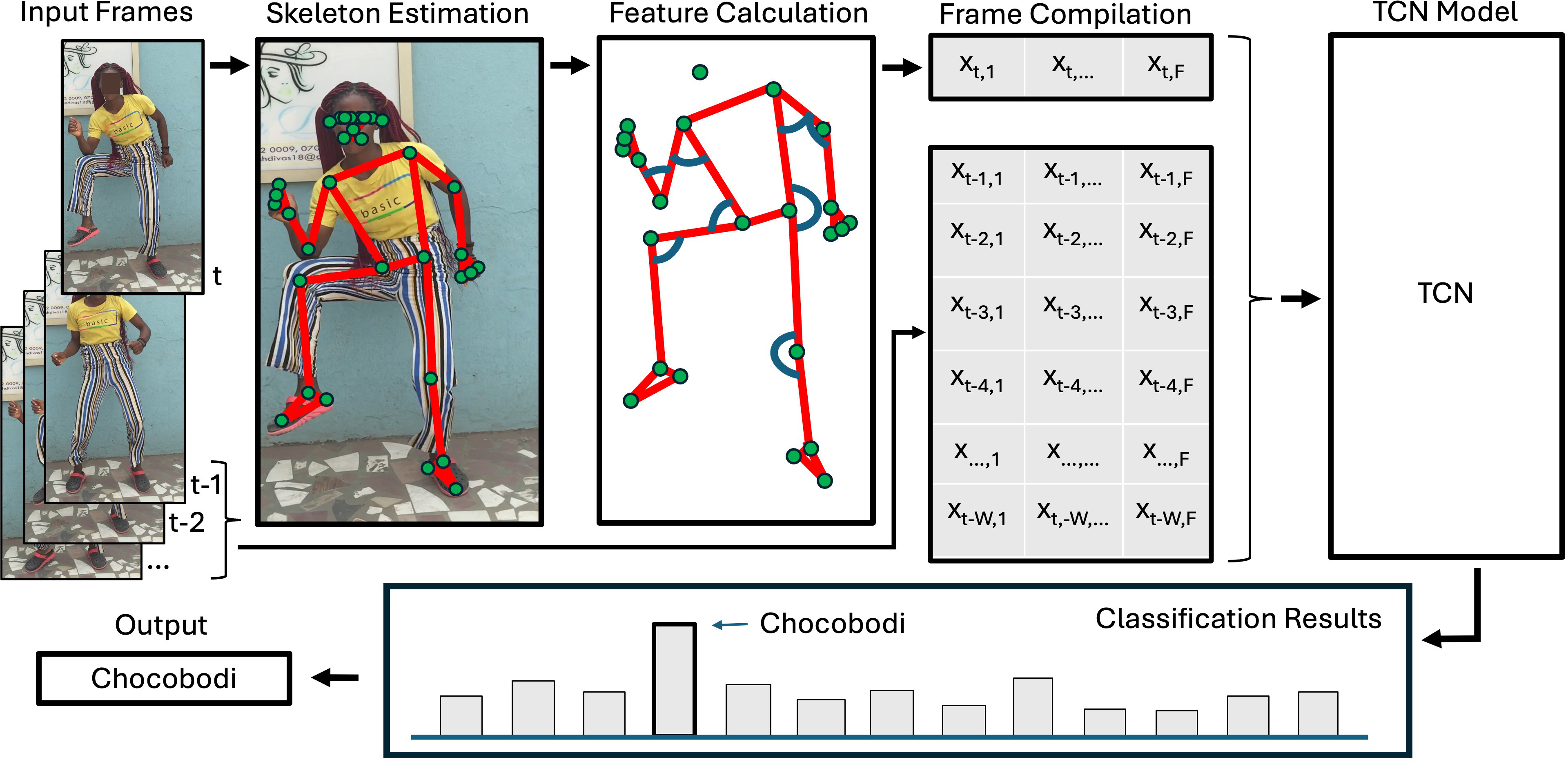}
    \caption{The dance move classification pipeline}
    \label{fig:pipeline}
\end{figure}

For example, in the case of MediaPipe, the human pose detection model estimates from each video frame the location of 33 body landmarks and calculates the following four values for each landmark: $x$ (the horizontal coordinate), $y$ (the vertical coordinate), $z$ (the depth of the point), and $visibility$ (with value in the range $[0-1]$ indicating the visibility status of the point). In our context, the 3-dimensional skeleton information in this dataset can enhance performance in human action recognition due to its rich informational content \cite{le2022comprehensive}. The locations of these 33 landmarks correspond to major joints in the human body as well as distinct facial landmarks and are shown in the skeleton estimation panel in Fig. ~\ref{fig:pipeline}. It should also be noted that another benefit of this data modality is that although it is very descriptive of the human pose, it does not contain any identifiable information of the human subjects in the video, such as facial characteristics, tattoos, or other information that could be considered sensitive, which satisfies the requirements of our institutionally approved data review protocol.

%\begin{figure}[h]
%  \centering
 % \includegraphics[width=0.5\textwidth]{figures/pose_landmarks_index.png}
%  \caption{Location of 33 Body Landmarks Detected by Mediapipe. Figure by Google. [Creative Commons 4.0]}
  %\url{https://developers.google.com/mediapipe/solutions/vision/pose_landmarker}
%  \Description{Image of the location of all 33 body landmarks detected by Mediapipe.}
 % \label{fig:landmark_locations}
%\end{figure}

%In Table ~\ref{tab:features_description} are the descriptions for the 5 features detected by Mediapipe.
%\begin{table}[]
%    \centering
%    \caption{Description of Mediapipe Features}
%    \label{tab:features_description}
%    \begin{tabular}{|c|c|}
%        \hline
%        Feature & Description \\
%        \hline
%        x & Normalized x-cord from [0, 1] \\
%        y & Normalized y-cord from [0, 1] \\
%        z & Depth of the point \\
%        presence & Confidence that point was detected \\
%        visibility & Likelihood that point was in frame \\
%        \hline
%    \end{tabular} 
%\end{table}

There are two possible forms of output provided by MediaPipe pose detection algorithm: normalized and world landmarks. The normalized landmarks are the projections of the 3D landmarks onto the image frame. For example, the normalized landmark coordinates $x=0$ and $y=0$ correspond to the top-left corner of the frame, while $x=1$ and $y=1$ correspond to the bottom-right corner of the frame, regardless of the aspect ratio. It should be noted that the values of $x$ or $y$ outside the $[0,1]$ range correspond to inferred locations that are not visible within the image frame. For the $z$ coordinate, the depth is measured relative to the center of the pelvis (midpoint between the hip) and normalized on a scale similar to that of the $x$ coordinate. Negative values of $z$ indicate that the landmark is closer to the camera than the pelvis, while values of $z>0$ indicate that it is farther away. In essence, the coordinate system for normalized landmarks is relative to the video frame and is useful for identifying the exact pixel location of each landmark.

World landmarks are 3D coordinate estimations in a relative 3D space based on the visual information captured in the video frames. More specifically, the coordinates of the world landmarks represent 3D points in a coordinate system with origin at the midpoint of the hips. However, it should be noted that due to the lack of geometric information in monocular systems, it is difficult to accurately estimate the actual real-world scale of the coordinate system, and consequently we cannot rely on these coordinates to make anthropometric measurements, such as estimating the real length of the limbs, without introducing additional input or constraints. However, regardless of the lack of scale, the world landmarks represent a reliable descriptor of the human pose, which satisfies our needs in this project. For the purposes of our experimentation, we trained our models using both normalized and world landmark features. 

For each video, all calculated features were stored in a raw feature array of size $(F, 33, 4)$, where $F$ denotes the number of frames in a video, 33 is the number of body landmarks, and 4 is the number of estimated values per landmark. For machine learning training or testing, this array can be sliced to retrieve the desired input based on the features selected for a particular experiment. For example, when training a model using only the $x$ and $y$ coordinates and with a 16-frame time window, the resulting size of the input array would be $(16, 33, 2)$. This array can then be reshaped into an array $X$ of size $(16, 66)$ for input into a machine learning model, where the first dimension is the temporal dimension and the second is the feature dimension.

\subsection{Additional feature Calculation}
\label{sec:additional_features}
In addition to the landmark position data extracted from the raw input video frames, other higher-order descriptors of the human posture can be calculated by utilizing the placement of each landmark/joint within the human skeleton and the relative dynamics between these joints \cite{song_2020}. For example, landmark velocities, joint angles, and angular velocities can be derived and included in the feature vector to test whether they can improve movement classification. 
 
Using the world landmark coordinates, we calculated angles between the major joints as additional features for the model. This step is illustrated in the feature calculation panel of Fig. ~\ref{fig:pipeline}. Table ~\ref{tab:body_angles} lists the calculated body angles and the corresponding points used to compute each angle as $\theta = arccos(\frac{u \cdot v}{|u||v|})$, where $u=p_1-p_2$ and $v=p_3-p_2$ are the vectors formed between the middle point ($p_2$) and the two other points ($p_1$ or $p_3$), respectively. The orientation of the torso was also calculated as two angles: between the $y$ axis and the medial vertical axis of the torso, and another between the $x$ axis and the shoulder line. In total, 10 angles were calculated and added to the feature vectors for model training. These angles were stored in a 2D array of size $(frames,10)$, so that it can be concatenated with the positional data along the feature axis to form a larger set of values for input into the model. 

\begin{table}[]
    \centering
    \begin{tabular}{|c|c|}
        \hline
        Body Angle & Landmarks Used: [$p_1$, $p_2$, $p_3$] \\
        \hline
        Right Elbow & [Right Shoulder, Right Elbow, Right Wrist] \\
        Left Elbow & [Left Shoulder, Left Elbow, Left Wrist] \\
        Right Shoulder & [Right Elbow, Right Shoulder, Right Hip] \\
        Left Shoulder & [Left Elbow, Left Shoulder, Left Hip] \\
        Right Hip & [Right Shoulder, Right Hip, Right Knee] \\
        Left Hip & [Left Shoulder, Left Hip, Left Knee] \\
        Right Knee & [Right Hip, Right Knee, Right Ankle] \\
        Left Knee & [Left Hip, Left Knee, Left Ankle] \\ 
        \hline
        \hline
        Torso X & [Right Shoulder, Left Shoulder],[1,0,0] \\ 
        Torso Y & [Shoulder Center, Hip Center],[0,1,0] \\ 
        \hline
    \end{tabular}
    \caption{Body angles calculated and the corresponding landmarks used}
    \label{tab:body_angles}
\end{table}

To provide additional input data to the model, velocities were calculated from both the positional and angular data by taking the inter-frame differences between corresponding values. This calculation nearly doubles the size of the input data, resulting in 2D arrays for velocity and angular velocity of size $(frames-1, features)$. These arrays were then padded with zeros to match the number of frames, allowing them to be concatenated along the features axis with the position and angle features for input into the model as illustrated in the frame compilation panel of Fig. ~\ref{fig:pipeline}.

The complete dataset of normalized and world landmark coordinates used as input in our experiments is available to download at \url{https://github.com/digitalworlds/ATUNDA/}, following the FAIR principles to ensure that the data is findable, accessible, interoperable, and reusable.

\section{Machine Learning Setup}
\label{sec:machine_learning}
After the preparation of the input feature data as described in Sec. \ref{sec:data_processing}, we used them to train a series of machine learning models for the classification of dance moves. This section describes the machine learning architecture and training parameters that were used, as well as the different configurations of the input feature vector chosen in each experiment.

\begin{figure}[h]
  \centering
  \includegraphics[width=1\textwidth]{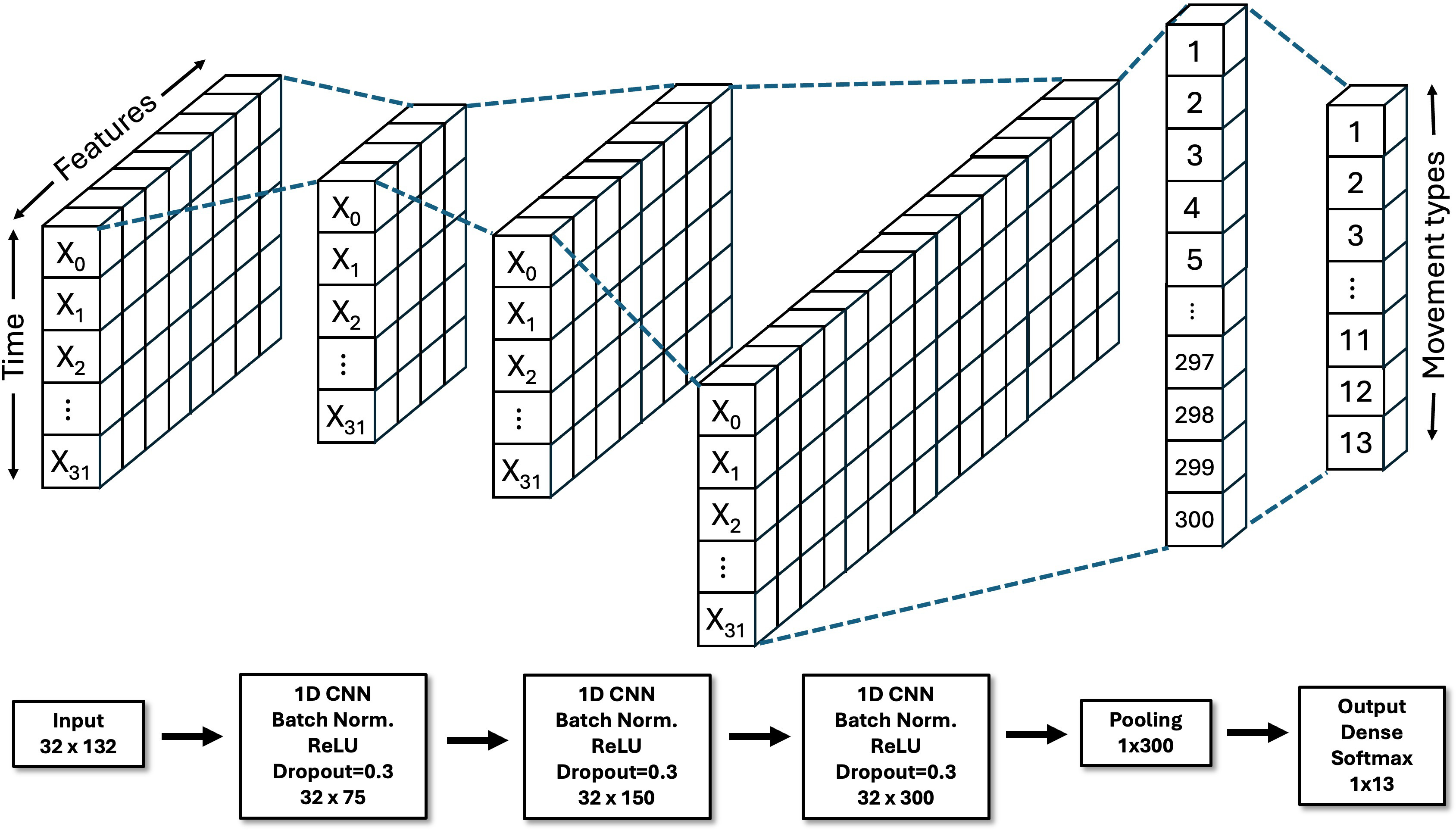}
  \caption{The TCN architecture with three convolutional layers used in our experiments. The exact number of features and input frames chosen for each experiment is reported in Tables \ref{tab:mediapipe_ablation} and \ref{tab:source_frame_comparison}.}
  \Description{Image of the model architecture used in all the models.}
  \label{fig:architecture}
\end{figure}

In all experiments we used a Temporal Convolutional Network (TCN) architecture with three convolutional blocks, as depicted in Fig. ~\ref{fig:architecture}. The input in each convolutional step is shown as a 2D array, with elements $x_{i,j}$, where the first dimension corresponds to the temporal domain (frames), while the second dimension is the number of input features per frame or in the case of the intermediate layers the number of filter outputs per frame. Each convolution step is followed by batch normalization, activation, and dropout. The final output is calculated after global average pooling and a dense layer as shown in Fig. \ref{fig:architecture}.

Configuring a temporal convolutional network requires selecting several key hyperparameters, such as the number of filters in each layer, the size of the convolution kernel, and the dilation rates. As shown by Ahmed, et al. \cite{inproceedings} increasing the number of filters can have a major impact on the classification accuracy of a model, at the cost of an increase in computational time. For our experiment, we utilized increasing filter sizes along the 3 convolutional blocks. After preliminary testing, our models performed best with 75, 150, and 300 filters across the 3 blocks. Similar studies \cite{TCN_LIDAR_2020, BaiTCN2018} found that a kernel of size $k=5$ and dilation rates of $d=1$, 2, and 4 across 3 TCN blocks were effective in TCN architectures, so we adopted these hyperparameters in our experiments.

With this configuration, the output of each convolution block is calculated as follows:
\begin{equation}
y_{t,f} = \sum_{i=1}^{5} \sum_{j=1}^{C_{\text{features}}} w^f_{i,j} \cdot x_{t - d \cdot i + p, \, j} + b^f
\end{equation}
where the first sum performs a convolution over the temporal domain of the input $x_{i,j}$ with a kernel of size $k=5$, while the second sum calculates a weighted sum of all the input features. The parameter $p$ corresponds to the padding at the beginning and the end of the input data along the temporal domain, and can be set according to a desired padding pattern. In our configuration we configured the padding so that the output has the same spatial dimensions as the input, also known as \texttt{padding='same'}. It should be noted that there is one weight matrix $w^f_{i,j}$ and one bias $b^f$ for each filter $f$ that contributes to the calculation of the respective output $y_{t,f}$. Therefore, 
according to the number of filters in our configuration, in the first, second, and third TCN blocks, there are 75, 150, and 300 sets of 2D weight matrices and bias parameters, respectively. The first dimension of the weight matrices corresponds to the receptive field of the convolutional block along the temporal domain. The effective receptive field size or effective kernel size is calculated as
\begin{equation}
k_{\text{eff}} = d \cdot (k - 1) + 1
\end{equation}
and increases with the dilation rate $d$. In our configuration, having a fixed kernel size and doubling the dilation rate in each convolutional block results in the following effective receptive field sizes and side paddings $p=(k_{\text{eff}}-1)/2$, as shown in Table \ref{configuration_table}.

\begin{table}[h]
\centering
\begin{tabular}{|c|c|c|c|c|}
\hline
\textbf{TCN block} & $d$ & $k_{\text{eff}}$ & \textbf{Total Padding} & $p$ \\
\hline
 1 & 1 &  5 & 4 & $2$ \\
2 & 2 & $9$  & $8$   & $4$ \\
3 & 4 & $17$ & $16$ & $8$ \\
\hline
\end{tabular}
\caption{Padding values for kernel size $k=5$ with different dilation rates using \texttt{padding='same'}}
\label{configuration_table}
\end{table}

 % The number of filters in the convolutional layer was chosen by performing a preliminary sequence of tests, in which we doubled the number of filters in each test and monitored the change in validation accuracy. The number of filters we tested was 128, 256, 512, 1024, 2048 respectively. Increasing the number of filters up to 1024 increased the validation accuracy. However a drop was observed when the number was further increased to 2048. For this reason the number of filters was set to 1024, for the main experiments presented in this paper.

Each convolutional block was followed by a Rectified Linear Unit (ReLU) as the activation function after batch normalization. Finally, a dropout layer completes each of the three structural blocks of our architecture. Using dropout layers for regularization can help increase robustness and the ability to generalize to new data \cite{electronics12143106}. Dropout layers have the hyperparameter of dropout rate, which affects how many neurons are turned off during training. For the purpose of human action recognition, Nan et al. \cite{nan2021comparison} found that a dropout rate around 0.2 was optimal, so in our preliminary tests we experimented with values $0.2\pm 0.1$. A dropout rate of 0.3 performed best and hence was chosen for our experiments.

\subsection{Model training}
\label{sec:model_training}
The architecture described in the previous section was employed in a series of experiments to train machine learning models and assess their ability to correctly classify the dance moves in the Atunda data bank. Each video was divided into fixed-length clips using a sliding temporal window, which was advanced by a stride to reduce unnecessary repetition of frames between clips. To maintain a consistent amount of overlap across clip lengths, the input stride was set to one quarter of the clip length, resulting in 75\% overlap between adjacent clips. To help prevent label bias caused by a lack of data in some types of dance moves, we evenly oversampled underrepresented movements so that each movement had the same number of clips in the training dataset.

In all experiments, we used the same seeded 5-fold cross-validation stratified by video, applying early stopping with a patience of 10 based on validation accuracy and retaining the best-performing weights. Because we used the same cross-validation split for each experiment, the results are meaningfully comparable between experiments. In order to avoid data leakage, all clips from a video were exclusively assigned to one set. Since we stratify only by video, some performers may appear in both the training and validation sets; however, many performers appear in multiple movement classes (in 5.41 classes on average), so the impact on data leakage is anticipated to be small, as the model cannot solely rely on characteristics of the performer to distinguish between movements.

We first trained 6 models using different input data configurations of the MediaPipe data as described in Table \ref{tab:mediapipe_ablation}. In each of the first six models, we introduced a different type of input data starting from normalized landmarks in Model 1, world landmarks in Model 2, body angles in Model 3, velocities and accelerations in Models 4 and 5, respectively, and used data augmentation in Model 6. 

Then, with Models 6-17 we compared different pose sources as we increase the clip size. In all of the previous models we used 32 frames as a baseline clip length, while in Models 6-9 we tested clip lengths of 32, 64, 96, and 128 frames, respectively. These models also used the full features (joint positions, angles, velocity, acceleration) and data augmentation from Model 6. Increasing the clip length provides more temporal information to the model and improves classification by smoothing out movement variability. However, as the clip length approaches the video length, the number of training clips per training video decreases, which can create an imbalance in the training data, as videos vary in length. We repeated these experiments using FMPose3D \cite{wang2026fmpose3dmonocular3dpose} for Models 10-13 and SAM4D \cite{gao2025sambody4d} for Models 14-17 as the pose sources to reduce our reliance on just the MediaPipe data and to compare how robust the different pose sources are at different clip lengths.

All 17 configurations were used to train TCN models with the architecture described in Fig. \ref{fig:architecture} using the Adam algorithm for optimization \cite{kingma2017adam}. All training and testing experiments were performed using a NVIDIA GB10 GPU on a DGX Spark with PyTorch library version 2.10.0 (NVIDIA CUDA 13.0, cuDNN 9.15.1), NumPy 1.26.4, SciPy 1.16.1, and scikit-learn 1.7.1 in Python 3.12.3.
The next section presents the results obtained from each model and discusses their performance in classifying dance moves. To facilitate the reproducibility of our experiments within an open science framework, the source code is available at \url{https://github.com/digitalworlds/ATUNDA/tree/main/code}.

\section{Results}

Table \ref{tab:mediapipe_ablation} summarizes the results of our MediaPipe input ablation in Models 1-6, and Table \ref{tab:source_frame_comparison} summarizes the results of our Pose-source and clip-length comparison in Models 6-17. The reported results were calculated by averaging the validation metrics across the 5-folds. The results indicate that the validation accuracy improved from 90.45\% in Model 1 to 96.86\% in Model 9 as a result of changing the type and size of the input feature vectors.

\begin{table}[]
    \centering
    \begin{tabular}{|c|c|c|c|c|c|c|}
        \hline
        Model & Frames & MediaPipe & Additional & Data & Validation & Macro F1 \\
        & & Landmarks & Features & Augmentation & Accuracy &  \\
        \hline
        1 & 32 & Normalized &  & No & 90.45 (1.21) & 90.29 (0.78) \\
        2 & 32 & World &  & No & 93.05 (1.86) & 93.04 (1.11) \\
        3 & 32 & World & Angles & No & 93.60 (1.80) & 93.84 (1.47) \\
        4 & 32 & World & Angles, Velocity & No & 94.40 (1.60) & 94.49 (1.89) \\
        5 & 32 & World & Angles, Velocity, Acceleration & No & 94.60 (1.80) & 94.63 (1.97) \\
        6 & 32 & World & Angles, Velocity, Acceleration & Yes & 94.54 (1.27) & 94.69 (1.48) \\
        \hline
    \end{tabular}
    \caption{MediaPipe input ablation results. Values are reported as mean (standard deviation) expressed as a percentage.}
    \label{tab:mediapipe_ablation}
\end{table}

More specifically, a comparison between Models 1 and 2 reveals that world landmarks (93.05\% validation accuracy) outperform normalized landmarks (90.45\% validation accuracy), which is expected given their superior ability to represent 3D pose independently of scale and perspective. This led us to choose world landmarks as the more suitable data input for the remaining MediaPipe Models (2-9).

In Models 3-5 we introduced the additional features angles, velocities, and accelerations as discussed in Sec. \ref{sec:additional_features}. The results show an incremental improvement of the validation accuracy of 1.11\% compared to performance without these additional features. Similarly, in Model 6 we introduced data augmentation as discussed in Sec. \ref{sec:machine_learning}. This produced comparable validation accuracy to Model 5 while slightly increasing Macro F1 and reducing the standard deviation.

Using Model 6 as our baseline, which outperformed all earlier configurations, we then focused on providing more temporal context to the TCN by increasing the input clip length, which had previously been fixed at 32 frames in all models up to Model 6. In Models 7–9, we increased the number of input frames from 32 up to 128 as shown in Table \ref{tab:source_frame_comparison}. This increase further improved the validation accuracy by 2.32\% to 96.86\%.

\begin{table}[]
    \centering
    \begin{tabular}{|c|c|c|c|c|c|}
        \hline
        Model & Pose Source & Frames & Stride & Validation Accuracy & Macro F1 \\
        \hline
        6 & MediaPipe & 32 & 8 & 94.54 (1.27) & 94.69 (1.48) \\
        7 & MediaPipe & 64 & 16 & 96.05 (1.32) & 96.22 (1.72) \\
        8 & MediaPipe & 96 & 24 & 96.11 (1.98) & 96.35 (2.41) \\
        9 & MediaPipe & 128 & 32 & 96.86 (2.02) & 96.68 (2.76) \\
        10 & FMPose3D & 32 & 8 & 90.71 (2.20) & 90.46 (1.61) \\
        11 & FMPose3D & 64 & 16 & 92.44 (2.58) & 92.04 (2.04) \\
        12 & FMPose3D & 96 & 24 & 93.82 (2.58) & 92.21 (2.98) \\
        13 & FMPose3D & 128 & 32 & 94.42 (2.73) & 94.07 (0.57) \\
        14 & SAM4D & 32 & 8 & 94.01 (2.44) & 93.09 (2.91) \\
        15 & SAM4D & 64 & 16 & 95.27 (2.65) & 94.08 (3.65) \\
        16 & SAM4D & 96 & 24 & 95.93 (2.96) & 94.20 (5.29) \\
        17 & SAM4D & 128 & 32 & 95.97 (3.12) & 93.49 (5.89) \\
        \hline
    \end{tabular}
    \caption{Pose-source and clip-length comparison. Joint positions, angles, velocity, acceleration, and data augmentation is used in all models. Values are reported as mean (standard deviation) expressed as a percentage.}
    \label{tab:source_frame_comparison}
\end{table}

\begin{figure}
    \centering
    \includegraphics[width=0.5\linewidth]{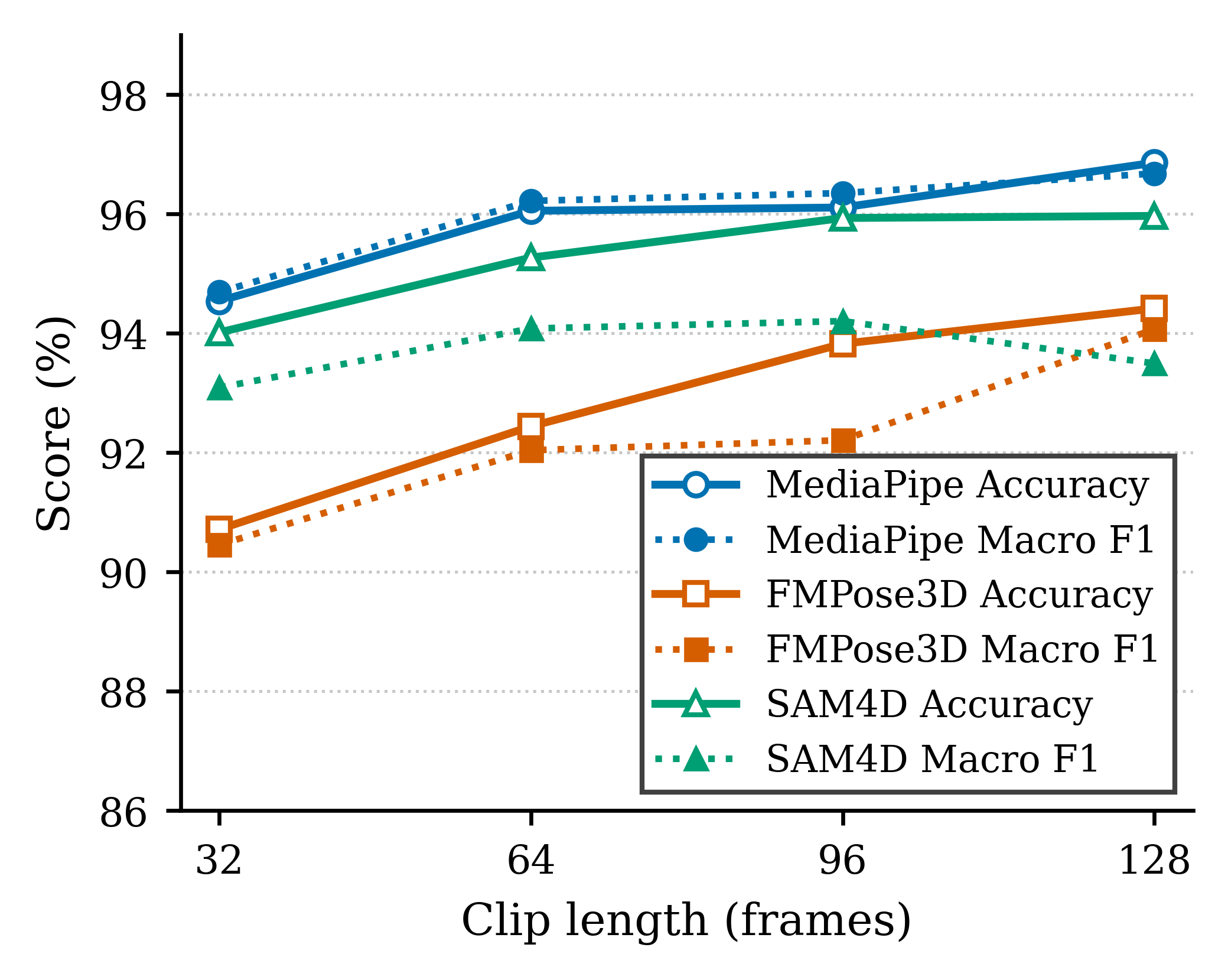}
    \caption{Plot of the accuracy and Macro F1 obtained from the three pose estimation methods from Table \ref{tab:source_frame_comparison} for various clip lengths.}
    \label{fig:source_frame_accuracy_macro_f1}
\end{figure}

% However, referring back to Model 1, the gap between validation and training accuracies was 6.86\%, indicating that the additional data from angles, velocities, frames, and augmentation helped reduce overfitting. Nevertheless, the dataset consisted of only 400 videos. The slight overfitting observed in Model 10 suggests that expanding the dataset with additional videos could further mitigate this issue.

We repeated the clip-length experiments of Models 6-9 Using FMPose3D \cite{wang2026fmpose3dmonocular3dpose} for Models 10-13, resulting in a 3.71\% accuracy increase up to 94.42\%, and repeated this again using SAM4D \cite{gao2025sambody4d} for Models 14-17, resulting in a 1.96\% increase up to 95.97\%. This shows FMPose3D performing much worse than MediaPipe, especially at lower clip lengths, and SAM4D had comparable but slightly worse performance across the clip lengths, with a notably worse Macro F1. Figure \ref{fig:source_frame_accuracy_macro_f1} shows the performance of  the three pose estimation methods from Table \ref{tab:source_frame_comparison} as a plot of the clip length.

Referring to Table~\ref{tab:video_description}, although 84 of the 397 videos (approximately 21\%) were recorded at 60 fps instead of 30 fps, the high model accuracy suggests that frame rate differences do not prevent effective classification. In other words, the models appear to distinguish movement types based on positions and positional changes rather than movement speed, allowing for some creative variation in the timing and tempo of the dance moves.

\begin{table}[ht]
  \centering
  \begin{tabular}{lcccc}
    \toprule
    \textbf{Class} & \textbf{Precision} & \textbf{Recall} & \textbf{F1-Score} & \textbf{Support} \\
    \midrule
    Akwaaba   & 1.00 & 0.97 & 0.99 & 75 \\
    Alanta    & 0.98 & 1.00 & 0.99 & 46 \\
    Bante     & 1.00 & 0.75 & 0.86 & 12 \\
    Chocobodi & 0.98 & 0.94 & 0.96 & 342 \\
    Etighi    & 0.99 & 0.99 & 0.99 & 143 \\
    Gbeku     & 0.95 & 0.95 & 0.95 & 63 \\
    Gwara     & 1.00 & 0.89 & 0.94 & 121 \\
    Jazzé     & 0.92 & 1.00 & 0.96 & 329 \\
    Obloblo   & 0.99 & 1.00 & 1.00 & 198 \\
    Omo Iya   & 0.96 & 0.96 & 0.96 & 47 \\
    Shooo     & 0.95 & 1.00 & 0.97 & 52 \\
    Swo       & 0.98 & 1.00 & 0.99 & 60 \\
    Wakanco   & 0.96 & 0.98 & 0.97 & 45 \\
    \midrule
    Accuracy     &      &      & 0.97 & 1533 \\
    Macro Avg    & 0.97 & 0.96 & 0.96 & 1533 \\
    Weighted Avg & 0.97 & 0.97 & 0.97 & 1533 \\
    \bottomrule
  \end{tabular}
  \caption{Per-class precision, recall, and F1-score for Model 9, the best-performing model.}
  \label{tab:classification_metrics_model9}
\end{table}

% \begin{figure}[h]
%     \centering
%     \includegraphics[width=0.5\textwidth]{figures/ROC.png}
%     \caption{Average ROC curves for all classes in Model 10}
%     \label{fig:roc}
% \end{figure}

% The ROC curve evaluates a model's ability to distinguish between positive and negative classes by comparing the true positive and false positive rates across different classification thresholds. While ROC curves are typically used for binary classification, they can be extended to multi-class models, as in this study. This is achieved by computing ROC metrics for each class individually, treating the target class as positive and all others as negative. The resulting ROC plots show that all 13 dance moves achieve a large area under the curve, indicating strong discriminative performance. 

A more detailed view of the classification performance for each dance move is provided in Table~\ref{tab:classification_metrics_model9},  which reports the precision, recall, and F1 scores per class. According to the table, most classes exhibit strong precision and recall, except Bante, which achieved a recall of only 0.75. This result is not unexpected, as Bante is the least represented movement in the dataset, referring back to Table~\ref{tab:video_description}, Bante only has 96.60 seconds of video. This difference in data is amplified at 128 frame clip lengths because the videos for this movement are shorter than others. Expanding the dataset with additional recordings of Bante could help reduce the bias of the model against Bante. 

% From this perspective, if the classifier is viewed as a similarity measure, the observed mislabeling suggests that bante and shooo are correctly identified as highly similar. This bias of shooo over bante is further supported by the precision of bante and recall of shooo. Both the precision of bante and the recall of shooo are 1.00 for model 9. Thus, every instance of shooo was correctly classified and every time bante was picked by the model it was correct. This supports the idea that the model was biased towards shooo so that every instance of shooo was classified correctly and only obvious instances of bante were chosen. 

\begin{figure}
    \centering
    \includegraphics[width=1\linewidth]{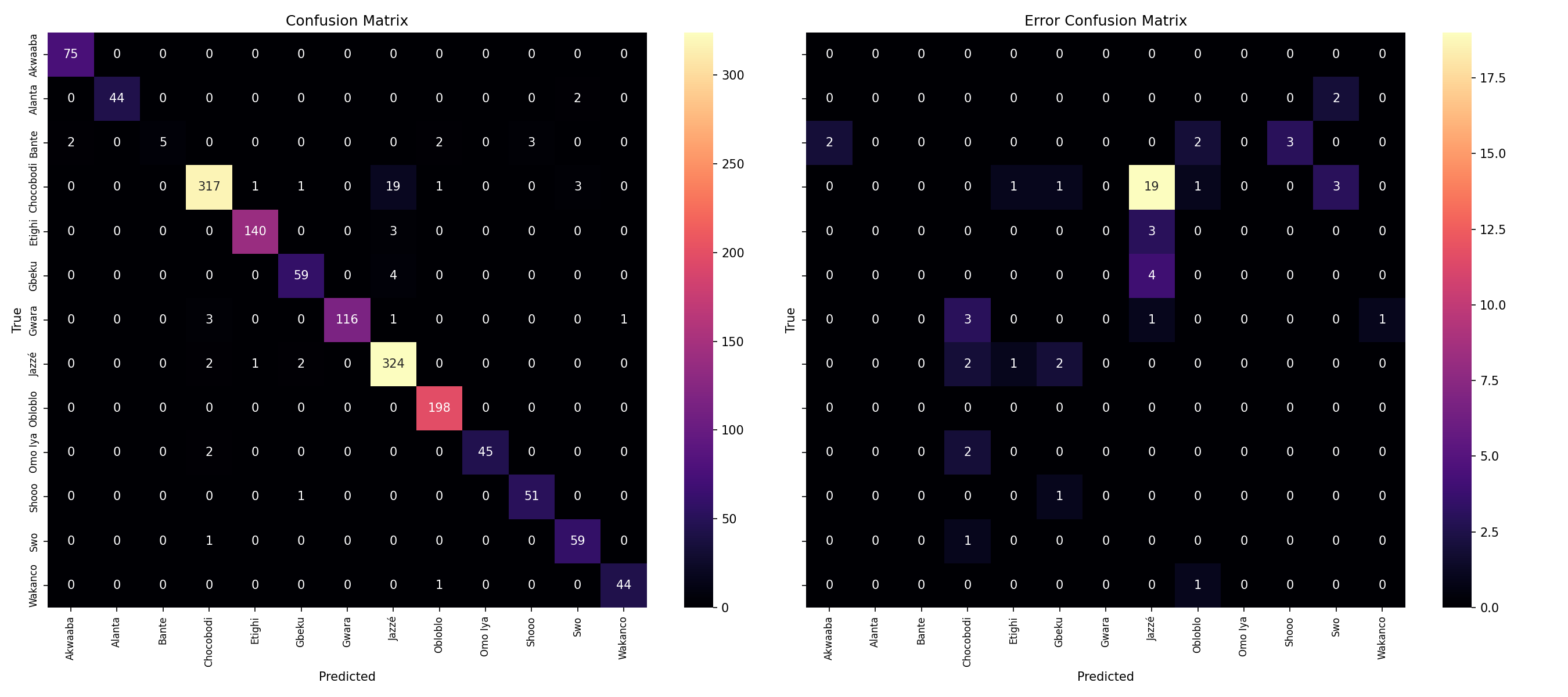}
    \caption{Confusion and Error Confusion Matrix for Model 9}
    \label{fig:cm_model9}
\end{figure}

In addition to the precision-recall table, we provide the confusion and error confusion matrices for Model 9 in Figure~\ref{fig:cm_model9}. The class with the most misclassifications was Chocobodi, as highlighted in the error confusion matrix. However, unlike Bante, Chocobodi still achieved a recall of 0.94, largely because it is the most represented class in the dataset, with 408.23 seconds of video—approximately four times more than Bante, as well as a mostly 60 fps frame rate rather than 30 fps. The misclassification of Chocobodi as Jazz\'e was by far the most common error. This was likely in part due to these two movements having the most data, and notably the only two movements mostly recorded at 60 fps. Therefore, despite the model achieving a validation accuracy of nearly 97\%, the observed misclassifications are likely influenced by class imbalance. Expanding the dataset, particularly for underrepresented classes, along with downsampling 60 fps videos, could help mitigate these errors and further improve overall performance.

Finally, we repeated the experiments using several baseline classification methods to evaluate the separability of the movements using non-deep-learning approaches. The baseline methods included: Majority (predicting the class with the largest number of samples as the simplest baseline), Linear Support Vector Machine (SVM), Radial Basis Function (RBF) SVM, and Random Forest. Unlike the TCN architecture, these methods are not specifically designed to capture temporal dependencies in sequential data. Therefore, a temporal summary representation of the feature vectors was computed and used as input to the baseline classifiers. The temporal summary consisted of four statistics computed for each feature dimension across 128 frames: mean, standard deviation, minimum, and maximum. These statistics were calculated from the same input configuration used in the previous experiments reported in Table~\ref{tab:source_frame_comparison}, i.e., joint positions, angles, velocity, and acceleration. This representation substantially reduced the feature dimensionality for the baseline methods and improved both accuracy and computation time in our experiments. Table~\ref{tab:classical_baselines} reports the validation accuracy and Macro F1 score for each combination of pose estimation and classification method. As expected, the performance of all baseline methods was inferior to that of the TCN. The best baseline performance was achieved by the Random Forest classifier using MediaPipe pose data, with an accuracy only 1.98 percentage points lower and a Macro F1 score only 4.23 percentage points below those of the TCN. These results indicate that the movements are reasonably distinguishable using only the relative landmark positions captured through temporal pose statistics. The Random Forest classifier appears to effectively exploit these feature patterns to separate the movement classes, particularly given the relatively small number of classes. We anticipate that this performance gap will increase as the number of classes expands in the second and third phases of our data collection efforts.

\begin{table}[]
    \centering
    \begin{tabular}{|c|c|c|c|}
        \hline
        Pose Source & Baseline & Validation Accuracy & Macro F1 \\
        \hline
        MediaPipe & Majority & 9.35 (1.56) & 1.34 (0.22) \\
        MediaPipe & Linear SVM & 89.53 (4.54) & 84.95 (5.87) \\
        MediaPipe & RBF SVM & 88.27 (4.00) & 84.64 (5.60) \\
        MediaPipe & Random Forest & 94.88 (2.22) & 92.45 (2.34) \\
        FMPose3D & Majority & 9.26 (1.41) & 1.34 (0.18) \\
        FMPose3D & Linear SVM & 79.42 (1.56) & 74.22 (1.81) \\
        FMPose3D & RBF SVM & 73.97 (4.04) & 71.22 (2.67) \\
        FMPose3D & Random Forest & 86.09 (1.48) & 81.37 (5.29) \\
        SAM4D & Majority & 9.13 (1.51) & 1.30 (0.17) \\
        SAM4D & Linear SVM & 91.03 (4.50) & 89.45 (6.36) \\
        SAM4D & RBF SVM & 88.48 (4.41) & 85.93 (5.04) \\
        SAM4D & Random Forest & 92.55 (3.13) & 88.13 (3.88) \\
        \hline
    \end{tabular}
    \caption{Classical machine learning baseline results for the final pose-source configurations from Table \ref{tab:source_frame_comparison}. Temporal summary (mean, standard deviation, minimum, and maximum) across 128 frames was used as input from the same feature vectors used in Table. \ref{tab:source_frame_comparison}. Values are reported as mean (standard deviation) expressed as a percentage.}
    \label{tab:classical_baselines}
\end{table}

\section{Discussion}
\label{sec:future_work}
This section discusses the results in the context of the broader goals of this project, outlines the limitations of the current study, and proposes future directions to address them.

This study represents an initial step toward our broader vision of developing a robust framework for the classification of Afrogenic dance movements and establishing digital metrics of similarity. Such metrics could eventually be used to track the evolution of dance styles, support proper attribution, and contribute to the wider discussion around the protection of the creative contributions of dance artists. As a first step, this work focused on compiling a novel dataset and establishing baseline performance metrics to guide future studies. Despite the considerable number of videos and movement types included, we consider this study pilot in nature within the context of our long-term goals. With the generous support of the Robert Wood Johnson Foundation, our interdisciplinary team is poised to expand the dataset by a factor of four by 2026. The data collection logistics and post-processing workflows developed in this pilot study have directly informed the design of the next phase of the project \cite{Pank2026Afrogenic}.

An important contribution of this work, not explicitly discussed elsewhere in the paper, is the development of the Atunda License, which enables creators to retain the rights to their performative videos contributed to the dataset. To design this license and ensure that the project promotes the ethical use of AI in the context of cultural heritage, we collaborated with experts in copyright law and user data rights. Users who license the Atunda dataset for AI model development or other research purposes are required to renew their license annually and re-download the dataset. If any contributors have removed their data in the interim, the updated dataset will reflect these changes upon renewal. By prioritizing creators’ control over their contributions, this approach ensures that data sharing advances AI research without compromising individual rights.

As discussed in the results section, one limitation arises from the dataset itself. The class distribution was highly imbalanced, with classes like bante containing significantly fewer samples than others, which may have contributed to its lower performance metrics. This issue could be mitigated by incorporating more data. Moving forward, one of our primary goals is to expand the dataset with additional videos to provide richer information to the models and achieve better class balance. In this regard, following the completion of this pilot study, the QDance Center has developed a public data collection platform to systematically crowdsource a larger dataset featuring more dance move categories  \cite{atundaLive}.

Furthermore, the architecture used in each of the deep-learning models was solely TCN-based. TCN architectures are known for capturing short-term dependencies, whereas recurrent-based architectures like LSTM's can better handle long term connections. As seen in our results, increasing frame sizes of the input led to significantly better accuracies. This accuracy could possibly be further improved by creating an ensemble of the TCN architecture with an LSTM to better handle dependencies established in longer frame inputs \cite{du2024tcnlstm}. Furthermore, the pose data for each frame were represented as a single vector. Restructuring the data into a spatial format based on the relative positions of body joints could potentially enable simultaneous convolution in both the spatial and temporal domains \cite{nan2021comparison}, a direction worth exploring.

While somewhat effective, the joint angle calculation was simplistic, as it computed only a single planar angle for each major joint. Incorporating additional angular information could better capture flexion-extension (bending) and pronation-supination (twisting) for each major joint in the upper and lower extremities, which could then be fed into the model. These additional angles would provide a more detailed representation of posture and could improve the model’s ability to distinguish between similar dance moves.

Despite these limitations, this paper, to the best of our knowledge, presented the first systematically collected AI-ready video dataset of Afrogenic dance moves. Furthermore, the baseline analysis presented in this paper can serve as a benchmark for evaluating future approaches to dance movement classification, particularly within the understudied field of African and African diaspora performative culture. By making the Atunda dataset available under the carefully designed Atunda License, we enable researchers to explore cutting-edge machine learning applications while ensuring that creators retain control over their performative works. We envision this dataset and its associated framework as a foundation for future developments, including automated attribution systems, digital metrics of similarity, and tools that protect artists’ intellectual and creative rights in the evolving digital landscape.  

\begin{acks}
The authors would like to thank the dance performers who participated in the data collection of the Atunda dataset. This project was funded in part by the Robert Wood Johnson Foundation, Grant number 79829. DC and TP would like to thank the University of Florida Center for Undergraduate Research and specifically the University Scholars Program for providing funding support for this project. AB would like to acknowledge that the NVIDIA DGX Spark was purchased with the funds from the University of Florida Research Foundation Professorship Award.  
\end{acks}

\bibliographystyle{apalike}
\bibliography{references}

\end{document}